\documentclass[11pt, article]{arxiv}
\usepackage[all]{hypcap}
\usepackage[comma,numbers,sort,compress]{natbib}
\usepackage{hyperref}[citecolor=magenta]
\usepackage[capitalize,noabbrev]{cleveref}
\hypersetup{
    colorlinks = true,
    citecolor = {StanfordRed},
    urlcolor=StanfordRed,
}

\usepackage{microtype}

\usepackage{subcaption}
\usepackage{booktabs} %

\usepackage{algorithm}
\usepackage{algorithmic}
\usepackage{mathrsfs}
\usepackage{setspace}

\usepackage{amsmath}
\usepackage{amssymb}
\usepackage{mathtools}
\usepackage{amsthm}
\usepackage[most,skins,theorems]{tcolorbox}
\tcbset{
  aibox/.style={
    width=\linewidth,
    top=7pt,
    bottom=2pt,
    colback=blue!6!white,
    colframe=black,
    colbacktitle=black,
    enhanced,
    center,
    attach boxed title to top left={yshift=-0.1in,xshift=0.15in},
    boxed title style={boxrule=0pt,colframe=white,},
  }
}
\newtcolorbox{AIbox}[2][]{aibox,title=#2,#1}

\usepackage{wrapfig}
\definecolor{lightblue}{rgb}{0.22,0.45,0.70}%
\usepackage{tabularx}

\definecolor{rliableolive}{HTML}{BBCC33}
\definecolor{rliableblue}{HTML}{77AADD}
\definecolor{rliablered}{HTML}{EE8866}

\usepackage{crossreftools}
\usepackage[suppress]{color-edits}

\usepackage{dsfont}
\usepackage{nicefrac}
\newtcolorbox{analysisbox}[1][]{
    enhanced jigsaw,
    colback=white,
    colframe=blue!75!black,
    fonttitle=\bfseries,
    boxsep=5pt,
    left=5pt,
    right=5pt,
    top=5pt,
    bottom=5pt,
    title=#1,
}
\definecolor{editInitialResponse}{RGB}{255, 235, 156} %
\definecolor{editBacktrack}{RGB}{0, 0, 139} %
\definecolor{editRevisedResponse}{RGB}{255, 182, 193} %

\usepackage{pifont}
\usepackage{soul}
\definecolor{highlightmistake}{RGB}{255, 179, 179} 
\definecolor{highlightcorrect}{RGB}{179, 255, 179}

\usepackage[capitalize,noabbrev]{cleveref}

\theoremstyle{plain}

\theoremstyle{definition}

\theoremstyle{remark}

\usepackage[textsize=tiny]{todonotes}

\usepackage{stfloats} 
\usepackage{multirow}
\usepackage{booktabs}
\usepackage{listings}
\usepackage{enumitem}

\usepackage{amsmath,amsfonts,bm}

\def\eqref#1{Eq.~\ref{#1}}

\def\1{\bm{1}}

\DeclareMathAlphabet{\mathsfit}{\encodingdefault}{\sfdefault}{m}{sl}
\SetMathAlphabet{\mathsfit}{bold}{\encodingdefault}{\sfdefault}{bx}{n}

\title{Temporal Preference Optimization of\\\vspace{0.2cm}Large Multimodal Models}

\author{Rui Li$^{*}$ \quad Xiaohan Wang$^{*}$  \quad  Yuhui Zhang  \quad 
Orr Zohar \quad Zeyu Wang \quad Serena Yeung-Levy \\
\vspace{0.5em}
Stanford University \\

\vspace{0.5em}
\texttt{Link: 
\href{https://ruili33.github.io/tpo_website}{Project Page}
~|~
\href{https://github.com/ruili33/TPO}{Code}
~|~
\href{https://huggingface.co/collections/ruili0/temporal-preference-optimization-67874b451f65db189fa35e10}{Dataset \& Checkpoints}}
}

\begin{document}

\maketitle
\begin{abstract}
\textbf{Abstract:} Despite recent advancements in video Large Multimodal Models (video-LMMs), accurate temporal grounding remains a key challenge. In this work, we introduce \textbf{Temporal} \textbf{Preference} \textbf{Optimization} (\textbf{TPO})—a post-training framework that unlocks superior temporal reasoning in video-LMMs without requiring human annotations.
TPO enables preference modeling by manipulating video inputs to generate contrastive responses, ensuring that preferred responses are more temporally grounded than dis-preferred ones. Through preference learning, TPO enhances the model’s capability to distinguish and localize events with better temporal reasoning.
Extensive experiments on LongVideoBench, MLVU, and Video-MME demonstrate that TPO significantly improves temporal grounding across multiple video-LMMs. Notably, LLaVA-Video-TPO achieves state-of-the-art performance among 7B models on Video-MME, establishing TPO as a scalable and effective solution for advancing temporal understanding in video analysis.
\end{abstract} 
\renewcommand\thefootnote{}\footnote{$^{*}$ Equal Contributions.}



\begin{figure}[ht]
    \centering
    \includegraphics[width=0.99\textwidth]{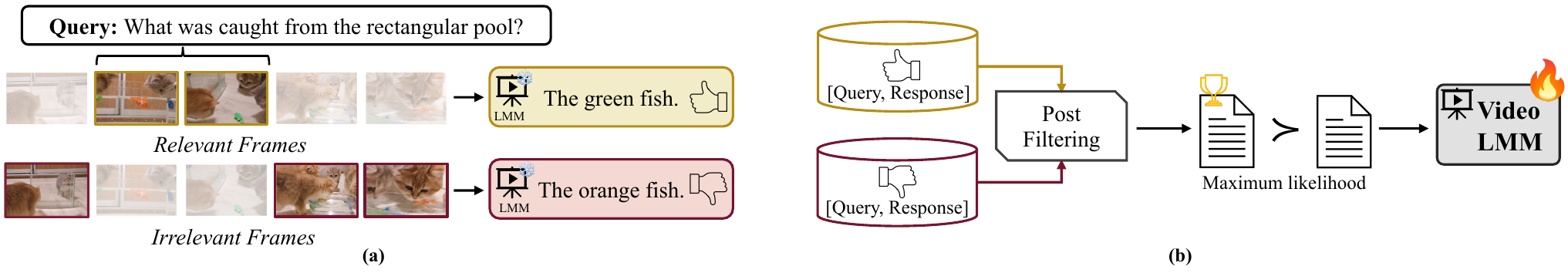}
    
    \caption{\textbf{Temporal Preference Optimization (TPO).} TPO is a post-training algorithm designed to enhance temporal comprehension in video-LMMs. In TPO, (a) preference data is generated by prompting the video-LMM with both well-grounded and manipulated (irrelevant or incomplete) video clips to collect contrastive response pairs. (b) These pairs undergo an LLM-based post-filtering process to remove noisy or misaligned samples. The generated preference data is then used in the preference optimization process, which improves the model’s temporal reasoning by prioritizing preferred responses, ultimately enhancing overall video understanding.}
    
    \label{fig:pull}
\end{figure}
\vspace{0.2in}
\abscontent
\section{Introduction}
\label{sec:intro}


Recent advancements in video Large Multimodal Models (video-LMMs) \cite{Qwen2VL,achiam2023gpt,reid2024gemini} have marked a significant step forward for generalizable video understanding. While image-based LMMs \cite{hong2024cogvlm2,Qwen-VL,lu2024deepseekvl} primarily focus on spatial reasoning, video-LMMs face the additional complexity of modeling temporal dependencies—a critical aspect for capturing the dynamic nature of video content. 


Most existing video-LMMs acquire temporal grounding implicitly during supervised finetuning by leveraging weak correspondences between input videos and textual responses \cite{chen2024sharegpt4video,zhang2024llavanextvideo}. While some responses may reference specific segments of a video, they lack explicit temporal alignment under next-token prediction training, limiting the model’s ability to learn precise temporal grounding.
Recently, alternative approaches \cite{ren2024timechat, chen2024timemarker, li2023videochat, huang2024lita,wang2024grounded} have emerged that incorporate explicit temporal annotations into training, enriching textual responses with structured temporal information as supervision. However, these methods rely on additional temporal annotations, which are costly to obtain and scale to large training datasets. 

In this work, we introduce Temporal Preference Optimization (TPO), a post-training framework designed to enhance the temporal grounding capabilities of video-LMMs. TPO systematically refines the pretrained video-LMMs' ability to distinguish temporally relevant content by leveraging contrastive responses from manipulated video inputs.
As shown in Fig.~\ref{fig:pull}, TPO first prompts a video-LMM with the same question on both the original and the corrupted video. Questions are formulated based on a set of video frames, with preferred responses generated using these frames. In contrast, dis-preferred responses are produced using the same question but paired with irrelevant or incomplete frame sets.
This pipeline ensures that preferred responses contain richer and more temporally relevant information than dis-preferred ones, thereby establishing a clear preference hierarchy.
By dynamically manipulating video inputs based on the query, TPO automatically injects temporal preferences into the preference dataset through simple input transformations. To further refine the dataset, a post-filtering process is applied to remove imperfect samples caused by errors from the pretrained video-LMMs and ambiguous preference data.
The resulting preference dataset is then used to optimize the model's temporal grounding capabilities via Direct Preference Optimization (DPO)~\cite{rafailov2024direct}, chosen for its flexibility and stability. This structured pipeline enables TPO to enhance the temporal reasoning capabilities of the base video-LMM using a curated post-training dataset, while preserving its pre-trained knowledge. This makes TPO a scalable and robust solution for advancing video understanding tasks.

We conducted extensive experiments on three challenging multimodal video understanding benchmarks, and the results clearly demonstrate that TPO significantly enhances the temporal grounding capabilities of video-LMMs. Specifically, TPO achieves performance gains of 2.9\% on LongVideoBench~\cite{wu2024longvideobench}, 3.1\% on MLVU~\cite{MLVU}, and 2.5\% on Video-MME~\cite{fu2024video}, when applied to the strong base model LongVA-7B~\cite{zhang2024long}. Furthermore, even when integrated with the state-of-the-art large-scale pretrained video-LMM, LLaVA-Video, TPO still delivers a 2.3\% improvement, establishing LLaVA-Video-TPO as the best-performing 7B model on the Video-MME benchmark.

\section{Preliminaries}
\label{sec:prelim}
\textbf{Preference learning} 
\cite{ouyang2022training,stiennon2020learning,ziegler2019fine} focuses on modeling human preferences to align model behavior with user expectations. In LLMs and image-LMMs, this involves training models to generate responses favored by users. This is typically achieved by collecting human feedback on pairs of model-generated outputs and learning a function that predicts which output is preferred. Formally, given an input x and two outputs $y^+$ (preferred) and $y^-$ (dispreferred), the model aims to satisfy:
\begin{equation}
    \pi_{\theta} (y^+|x)>\pi_{\theta} (y^-|x)
\end{equation}
where $\pi_{\theta}(y|x)$ is the model's probability of generating output $y$ given input $x$ with parameters $\theta$.
In the context of video-LMMs, a preference dataset $\mathcal{D}$ is constructed as a collection of tuples $(V, q, r^+, r^-)$, where $V$ denotes a video, $q$ represents a query, $r^+$ is the preferred temporally grounded response, and $r^-$ is the dispreferred response. 

\textbf{Direct Preference Optimization} (DPO) \cite{rafailov2024direct} is a methodology that directly integrates human preference data into the optimization of model parameters. Compared to Proximal Policy Optimization  (PPO) \cite{ouyang2022training,stiennon2020learning,ziegler2019fine}, another popular preference learning implementation,  DPO eliminates the need for explicit reward models or complex reinforcement learning algorithms. By leveraging human preference data as a guiding signal during optimization, DPO enhances the model’s ability to generate outputs that are better aligned with human values and expectations.
\section{Temporal Preference Optimization}
While prior works focus primarily on aligning LLM outputs with human preferences, our approach uniquely aligns model outputs with intrinsic temporal preferences in videos. 
To achieve this, we propose Temporal Preference Optimization (TPO), illustrated in Fig.~\ref{fig:method}, which significantly enhances the video reasoning capabilities of video-LMMs. 
TPO systematically incorporates temporal grounding into the optimization process through creating preference  pairs from  the contrast between meticulously manipulated video inputs (Sec.~\ref{section:preference_localized}).
To further enhance the quality of these preference data, we introduce a rule-based post-filtering step (Sec. \ref{section:post_filtering}). Finally, Direct Preference Optimization (Sec. \ref{section:preference_optimization}) is leveraged to optimize the model towards temporally preferred outputs without compromising its original pretrained capabilities.

\subsection{Temporal Preference Modeling}
\label{section:preference_localized}
\paragraph{Query Generation.}
Given a video $\bf{V}$, we first sample a segment containing a set of frames $S_a$, which may be a subset of the video or the entire sequence of frames. To generate descriptive context, we employ an image-based vision-language model (CogVLM2 \cite{hong2024cogvlm2}) to generate captions for each frame in $S_a$. These captions serve as the foundation for constructing targeted questions.
To ensure diversity and relevance, we design multiple question types and use a structured question-generation prompt, as shown in Fig. \ref{fig:prompt-query} (Appendix), incorporating the generated captions. This prompt is then processed by a large language model (GPT-4o-mini) to produce a set of candidate questions specifically tailored to the sampled video frames, resulting in a set of questions $S_q$. This approach ensures that the generated questions are  contextually relevant that allows precise control over subsequent response generation.


\begin{figure*}[ht]
    \centering
    \includegraphics[width=1\linewidth]{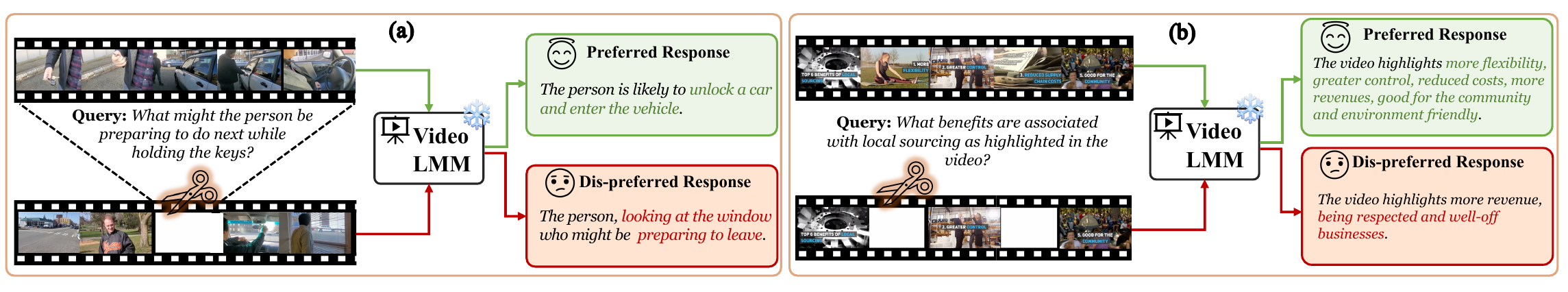}
    \caption{In TPO, we introduce two approaches for temporal preference data generation. Preferred responses are generated using questions and their corresponding frames for strong temporal grounding. For dis-preferred responses, we introduce:
    (a) \textbf{Generation with Irrelevant Information}, where all relevant frames are excluded. (b) \textbf{Generation with Incomplete Information}, where only a partial subset of relevant frames is used.
    These manipulated clips create contrastive response pairs, highlighting differences between well-grounded and manipulated video content. This contrast serves as a learning signal to enhance the model’s temporal reasoning.}
    \label{fig:method}
\end{figure*}
\paragraph{Preferred Response Generation.} 
Preferred responses in the curated dataset are expected to be strongly grounded in the corresponding temporal content. To achieve this, we use the question set $S_q$ along with their corresponding frame set $S_a$ as input to the video-LMM. By ensuring that the provided video frames are highly relevant to the query, we create conditions that maximize the likelihood of the model generating a high-quality, temporally grounded response. This process guarantees that the preferred responses align with the ideal characteristics for effective temporal grounding in video-LMMs.

\paragraph{Dis-Preferred Response Generation.}
The dis-preferred responses in the preference dataset represent the type of outputs the model aims to discourage, specifically those where it fails to localize relevant information in the video. These responses expose shortcomings in temporal reasoning by highlighting cases where the model struggles to align its predictions with the actual video content.
To generate these dis-preferred responses, we manipulate the video inputs to simulate imperfect temporal grounding. As illustrated in Fig.~\ref{fig:method}, we introduce two strategies for constructing the input frame set $S_b$ used in dis-preferred response generation:

\textbf{(a) Generation with Irrelevant Information}: To simulate an extreme failure case where the model misses all relevant frames, we construct $S_b$ by excluding the relevant frame set $S_a$ and instead sampling from the remaining frames of the video. This ensures that $S_b$ contains only irrelevant content, forcing the model to generate a response based on unrelated visual information.

\textbf{(b) Generation with Incomplete Information}: To mimic scenarios where the model has access to only partial relevant information, we construct $S_b$ by randomly sampling a subset of $S_a$. This setup introduces gaps in the temporal context, making it harder for the model to fully comprehend the event described in the query.

Unlike preferred responses, which are grounded in fully relevant video segments, these manipulated setups significantly increase ambiguity and noise by partially or completely omitting critical visual content. As a result, the model is forced to rely on incomplete or misleading information, making temporal reasoning errors and hallucinations more likely. This intentional contrast between preferred and dis-preferred responses serves as a strong learning signal, helping refine the model’s ability to distinguish and accurately localize events in time, ultimately enhancing its temporal reasoning capabilities.

\subsection{LLM-based Post-Filtering}
\label{section:post_filtering}
Although we design the preferred responses to be of higher quality than the dis-preferred responses, this distinction is not always guaranteed due to the imperfect video understanding capabilities of the base video-LMMs. In some cases, errors in response generation may lead to misaligned preference pairs, where the preferred response contains noise or the dis-preferred response is of better quality than expected.

To enhance data quality and reduce noise, we introduce a post-filtering pipeline leveraging an LLM (GPT-4o-mini). Specifically, we provide the model with the key frame captions of $S_a$, along with their corresponding queries and preference data pairs, and instruct it to filter out samples that meet predefined criteria (detailed prompts are shown in Fig. \ref{fig:prompt-filter1} in the Appendix). The filtering rules target cases where:
1) The dis-preferred response is of higher quality than the preferred response.
2) The preferred response is factually incorrect or misaligned with the video content.
3) The query itself is ambiguous, making preference ranking unreliable.
By incorporating this post-filtering step, we effectively eliminate problematic cases that could introduce noise into training, resulting in a refined, higher-quality dataset that better supports effective model optimization and improves temporal grounding performance.

\subsection{Training Objective}
\label{section:preference_optimization}
The generated preference dataset is leveraged to optimize the temporal grounding capabilities of video-LMMs using Direct Preference Optimization (DPO)~\cite{rafailov2024direct}, selected for its robustness and effectiveness in preference-based learning.

 Given the preference dataset D $(V, q, r^+, r^-)$ and the video-LMM $\pi_{\theta}$, the DPO loss function is defined as:
\begin{align}
    &L_{DPO}(\pi_{\theta};\pi_{ref})=-E_{(V,q,r^+, r^-)\sim \mathcal{D}} \nonumber\\ &\left[ log\sigma(\beta(\log \frac{\pi_{\theta}(r^+|V,q)}{\pi_{ref}(r^+|V,q)}
    -\log \frac{\pi_{\theta}(r^-|V,q)}{\pi_{ref}(r^-|V,q)}))\right] 
\end{align}
where $\sigma$ is the sigmoid function. This objective drives the model to assign higher probabilities to preferred outputs, aligning its behavior more closely with human judgments, while preventing the model from deviating too much from its pretrained distribution.

To better align the model with the preferred responses, we incorporate a supervised fine-tuning objective into the DPO training framework. This combined objective is controlled by the hyperparameter $\alpha$, following \cite{chen2021large,deng2024enhancing,chen2023understanding}.
 \begin{equation}
     L_{SFT}(\pi_{\theta})=-E_{(V,q,r^+, r^-)\sim \mathcal{D}} \log \pi_{\theta}(r^+|V,q)
 \end{equation}
 \begin{align}
     L(\pi_{\theta};\pi_{ref})=L_{DPO}+\alpha L_{SFT}
 \end{align}
\section{Experiments}
\subsection{Experimental Settings}

\paragraph{Evaluation Benchmarks}
We evaluate TPO and baselines on three widely recognized benchmarks in multimodal video understanding.
\begin{itemize}
    \item \textbf{Video-MME} \cite{fu2024video} offers a comprehensive multi-modal evaluation across diverse video lengths, spanning from 11 seconds to 1 hour.
    \item \textbf{LongVideoBench} \cite{wu2024longvideobench} emphasizes reasoning tasks within extended video contexts.
    \item \textbf{MLVU} \cite{MLVU} supports multitask evaluation specifically designed for long-form video understanding. 
\end{itemize}

\begin{table*}[t]
\centering
\begin{tabular}{lccccccc}
\toprule
\multirow{2}{*}{\textbf{Model}} & \multirow{2}{*}{\textbf{Size}} & \multirow{2}{*}{\textbf{LongVideo}} & \multirow{2}{*}{\textbf{MLVU }} & \multicolumn{4}{c}{\textbf{Video-MME}} \\
\cmidrule(lr){5-8}
 &  & \textbf{Bench} &  \textbf{(M-avg)}& \textbf{Short} & \textbf{Medium} & \textbf{Long} & \textbf{Average} \\
\midrule
GPT-4o \cite{achiam2023gpt} & - & 66.7 & 64.6 & 80.0$_{82.8}$ & 70.3$_{76.6}$ & 65.3$_{72.1}$ & 71.9$_{77.2}$ \\
\midrule
Video-LLaVA \cite{lin2023video} & 7B & 39.1 & 47.3 & 45.3$_{46.1}$ & 38.0$_{40.7}$ & 36.2$_{38.1}$ & 39.9$_{41.6}$ \\
LLaVA-1.5 \cite{liu2023llava}& 7B&40.3&-&-&-&-&-\\
PLLaVA \cite{xu2024pllava}&7B&40.2&-&-&-&-&-\\
Qwen-VL-Max \cite{Qwen-VL}&-&-&42.2&55.8$_{57.6}$&49.2$_{48.9}$&48.9$_{47.0}$&51.3$_{51.2}$\\
ShareGPT4Video \cite{chen2024sharegpt4video} & 8B & 39.7 & 46.4 & 48.3$_{53.6}$ & 36.3$_{39.3}$ & 35.0$_{37.9}$ & 39.9$_{43.6}$ \\
InternVL-Chat-V1.5 \cite{chen2023internvl} & 20B & 51.2 & 50.4 & 50.7$_{52.4}$  & 60.2$_{61.7}$  & 46.4$_{49.1}$  & 45.6$_{46.6}$  \\
VideoChat2 \cite{2023videochat} & 7B & 39.3 & 47.9 & 48.3$_{52.8}$  & 37.0$_{39.4}$  & 33.2$_{39.2}$  & 39.5$_{43.8}$  \\
LongLLaVA  \cite{long-llava-qwen2-7b-2024}& 7B & - & 56.3 & 61.9$_{66.2}$  & 51.4$_{54.7}$ & 45.4$_{50.3}$ & 52.9$_{57.1}$ \\
Video-CCAM \cite{fei2024video}&14B&-&63.1&62.2$_{66.0}$ &50.6$_{56.3}$ &46.7$_{49.9}$ &53.2$_{57.4}$ \\
NVILA \cite{liu2024nvila} & 7B & 57.7&70.1&75.7$_{77.6}$ &62.2$_{69.0}$ &54.8$_{63.3}$ &64.2$_{70.0}$ \\
Qwen2-VL \cite{Qwen2VL} & 7B & 55.6&-&-&-&-&63.3$_{69.0}$\\
Apollo \cite{apollo} & 7B & 58.5&70.9&-&-&-&61.3$_{63.3}$\\
\midrule
LongVA-7B \cite{zhang2024long} & 7B & 51.3 & 58.8 & 61.1$_{61.6}$  & 50.4$_{53.6}$  & 46.2$_{47.6}$  & 52.6$_{54.3}$  \\
LLaVA-Video-7B \cite{zhang2024videoinstructiontuningsynthetic} & 7B & 58.2 & 70.8 & - & - & - & 63.3$_{69.7}$\\
\midrule

\textbf{LongVA-TPO}~\footnotesize{(\textbf{\texttt{ours}})} & 7B & 54.2 & 61.7 & 63.1$_{66.6}$ & 54.8$_{55.3}$ & 47.4$_{47.9}$ & 55.1$_{56.6}$ \\
\textbf{LLaVA-Video-TPO}~\footnotesize{(\textbf{\texttt{ours}})} & 7B & \textbf{60.1} & \textbf{71.1} & \textbf{76.8}$_{\textbf{78.7}}$ & \textbf{64.6}$_{\textbf{69.4}}$ & \textbf{55.4}$_{\textbf{66.4}}$ & \textbf{65.6}$_{\textbf{71.5}}$ \\
\bottomrule
\end{tabular}
\caption{Results on LongVideoBench \cite{wu2024longvideobench}, MLVU \cite{MLVU} and Video-MME \cite{fu2024video} compared with state-of-the-art models. The Video-MME results are presented in the format w/o subs$_{ w/ subs}$.}
\label{tab:results}
\end{table*}

\begin{table*}[t]
\centering
\begin{tabular}{lcccccc}
\toprule
\multirow{2}{*}{\textbf{Model}} & \multirow{2}{*}{\textbf{LongVideoBench}} & \multirow{2}{*}{\textbf{MLVU}} & \multicolumn{4}{c}{\textbf{Video-MME}} \\
\cmidrule(lr){4-7}
 &  & \textbf{(M-avg)} & \textbf{Short} & \textbf{Medium} & \textbf{Long} & \textbf{Average} \\
\midrule
LongVA-7B \cite{zhang2024long} & 51.3 & 58.8 & 61.1$_{61.6}$ & 50.4$_{53.6}$ & 46.2$_{47.6}$ & 52.6$_{54.3}$ \\
$~~\textbf{+}$ SFT$_{\text{Self}}$ &52.7&58.9&62.6$_{\textbf{67.7}}$&52.4$_{52.7}$&46.8$_{47.4}$&53.9$_{55.9}$\\
$~~\textbf{+}$ SFT$_{\text{LLM}}$ &53.1&59.9&\textbf{63.7}$_{64.9}$&52.6$_{54.3}$& 46.3$_{47.9}$& 54.2$_{55.7}$\\
$~~\textbf{+}$ Hound-DPO$^\dagger$ ~\cite{zhang2024direct,zhang2024long}&52.8&59.1&62.2$_{65.8}$&52.4$_{54.8}$&46.1$_{46.3}$&53.6$_{55.6}$\\
$~~\textbf{+}$ Hound-DPO$^*$ ~\cite{zhang2024direct,zhang2024long} &52.6&59.3&63.1$_{65.9}$&50.8$_{54.7}$&47.2$_{47.0}$&53.7$_{55.9}$\\
\midrule
\textbf{LongVA-TPO}~\footnotesize{(\textbf{\texttt{ours}})} & \textbf{54.2} & \textbf{61.7} & 63.1$_{66.6}$ & \textbf{54.8}$_{\textbf{55.3}}$ & \textbf{47.4}$_{\textbf{47.9}}$ & \textbf{55.1}$_{\textbf{56.6}}$ \\
\bottomrule
\end{tabular}
\caption{Results of LongVA-TPO on LongVideoBench \cite{wu2024longvideobench}, MLVU \cite{MLVU} and Video-MME \cite{fu2024video} benchmarks compared to 3 baseline methods mentioned in \ref{section:results}. The Video-MME results are presented in the format ``w/o subs / w/ subs". The results for LongVA and LongVA+Hound-DPO$^\dagger$ are based on publicly available checkpoints, and for LongVA+Hound-DPO$^*$ are based on our implementation on our collected video datasets, while the other results are evaluated using our trained model.}
\label{tab:results_comparison}
\end{table*}

\paragraph{Models}
We test the effectiveness of TPO on two popular video-LMMs, LongVA-7B \cite{zhang2024long} and LLaVA-Video-7B \cite{zhang2024videoinstructiontuningsynthetic}, deriving the following models:
\begin{enumerate}
    \item LongVA-TPO: optimized based on LongVA-7B \cite{zhang2024long}, a capable video-LMM with the long-context video understanding capability transferred from language.
    \item LLaVA-Video-TPO: optimized based on LLaVA-Video-7B \cite{zhang2024videoinstructiontuningsynthetic}, the current state-of-the-art 7B video-LMM. 
\end{enumerate}

Without other states, our ablation study and analysis utilize LongVA-TPO by default.
\paragraph{Implementation Details} 
For the video source 
of preference dataset generation, we manually curated 200 keywords, which we used to retrieve 8k videos from the internet to curate a diverse and comprehensive dataset. 
From these crawled videos, we created 10k preference data pairs for LongVA-TPO using our established pipeline. For LLaVA-Video-TPO, we employ a subset of the original LLaVA-Video-178K dataset, which was used for supervised fine-tuning (SFT), to generate TPO data, resulting in a total of 10K preference data pairs.

The model is trained on 8 Nvidia A100 80GB GPUs, with a batch size of 64. For the preference optimization on LongVA, we set the KL-divergence weight $\beta=0.3$ and the SFT loss weight $\alpha=0.5$, while for LLaVA-Video, we set the KL-divergence weight $\beta=0.2$ and the SFT loss weight $\alpha=1$. We train the model  on our curated data for 1 epoch. It takes about 4 hours for TPO to perform on LongVA-7B with a learning rate of $4e^{-6}$ and also about 4 hours for LLaVA-Video-7B with a learning rate of $3e^{-7}$. During data preparation, we employ the GPT-4o-mini language model (text-only input) for question curation and post-filtering. This choice balances cost-effectiveness with efficiency, facilitating a streamlined and scalable data processing workflow.

\subsection{Results}
\label{section:results}
We conducted comprehensive experiments across three established datasets to rigorously assess the effectiveness of TPO in long-form video understanding tasks.
We first compare TPO with three different training strategies:
\begin{itemize}
    \item SFT$_{\text{Self}}$: Supervised fine-tuning using the self-generated data. For a fair comparison, we utilize the same preferred response in our curated preference dataset to optimize LongVA.
    \item SFT$_{\text{LLM}}$: Supervised fine-tuning using the LLM-generated data. Following the commonly used data curation pipeline~\cite {chen2024sharegpt4video,zhang2024llavanextvideo}. We employ LLM (GPT-4o-mini) to generate responses given the query and the video captions, which are subsequently used to perform supervised fine-tuning on LongVA. We use the same video data as TPO for fair comparison.
    \item Hound-DPO~\cite{zhang2024direct} is a previous method that applies Direct Preference Optimization (DPO)~\cite{rafailov2024direct} on video-LMMs. Their approach leverages ChatGPT~\cite{achiam2023gpt} to generate ratings for preference data, resulting in a dataset of 17k samples. In contrast, TPO employs a smaller preference dataset generated through a self-generation pipeline, offering a more streamlined alternative. Besides, to ablate the data source's effect, we also implement Hound-DPO based on our collected dataset with the same data scale.    
\end{itemize}
The primary experimental results, presented in Table \ref{tab:results_comparison}, compare TPO against the baseline methods on LongVA. The results consistently indicate that LongVA-TPO achieves superior performance, with improvements of 2.9\%, 3.1\%, and 2.5\% on LongVideoBench \cite{wu2024longvideobench}, MLVU \cite{MLVU}, and Video-MME \cite{fu2024video}, respectively. These findings underscore TPO's capacity to enhance the general video understanding capabilities of a pre-trained video-LMM.

Compared to SFT$_{\text{Self}}$, LongVA-TPO achieves a consistent performance gain of 1.2\% to 2.8\% by utilizing a carefully designed temporal dis-preferred response to contrast with the preferred response. Furthermore, LongVA-TPO outperforms SFT$_{\text{LLM}}$, demonstrating the effectiveness and stability of our self-training paradigm. When compared to Hound-DPO \cite{zhang2024direct}, LongVA-TPO achieves a significant performance improvement by injecting temporal preference priors into the dataset. However, LongVA-TPO underperforms SFT methods on the Video-MME-short subset, which is expected since LongVA-TPO primarily focuses on optimizing temporal reasoning for video understanding.

In addition, the comparisons between TPO and current state-of-the-art video-LMMs are presented in Table \ref{tab:results}. With the introduction of TPO, both the LongVA-TPO and LLaVA-Video-TPO models significantly outperform their corresponding baselines  2.5\% and 2.3\% on the video-MME benchmark, demonstrating the efficacy of our TPO pipeline.
After TPO on LLaVA-Video-7B, our LLaVA-Video-TPO model outperforms all 16 baseline models in the table, including the concurrent work, NVILA \cite{liu2024nvila}, as well as several 14B and 20B models, achieving state-of-the-art results on video understanding. The original LongVA model performed worse than Video-CCAM \cite{fei2024video} and LongLLaVA \cite{long-llava-qwen2-7b-2024} on the Video-MME benchmark. However, after incorporating TPO, it successfully outperformed these competitive baselines on Video-MME. 
Overall, LLaVA-Video-TPO achieves the strongest 7B model on Video-MME, setting a new state-of-the-art performance on video comprehension.

\begin{figure*}[htbp]
    \centering
    \begin{minipage}[t]{0.99\textwidth}
        \centering
        \includegraphics[width=\textwidth]{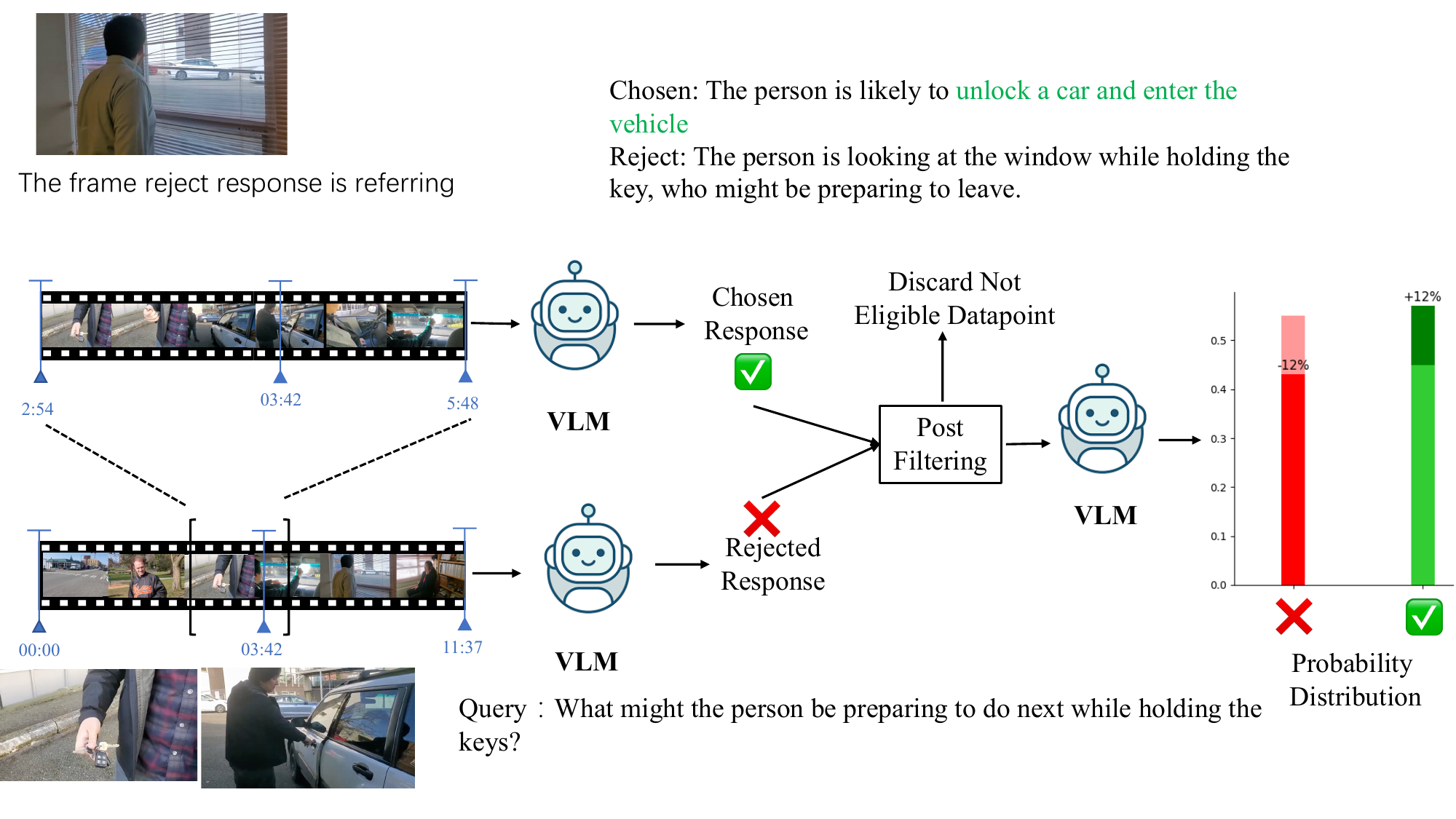}
    \end{minipage}

    \caption{Qualitative comparison between LongVA-TPO model and LongVA on two videos from VideoMME benchmark. 
    }
        \label{fig:qualitative}
\end{figure*}


\subsection{Ablation Study}
\subsubsection{Performance with Different Input Frame Count}
We evaluate the performance of both our LongVA-TPO model and the original LongVA model across varying input lengths, ranging from 16 to 128 frames, as shown in Fig.~\ref{fig:num_frame}. The results indicate that the LongVA model experiences performance degradation with 128 frames compared to 64 frames. In contrast, our LongVA-TPO model consistently benefits from longer input sequences, leveraging the additional information effectively. This demonstrates the LongVA-TPO model’s robustness in handling extended inputs and its capacity to localize relevant information within long sequences, further validating the efficacy of TPO.
\begin{figure*}[htbp]
    \centering
    \begin{minipage}[t]{\textwidth}
        \centering
        \includegraphics[width=\textwidth]{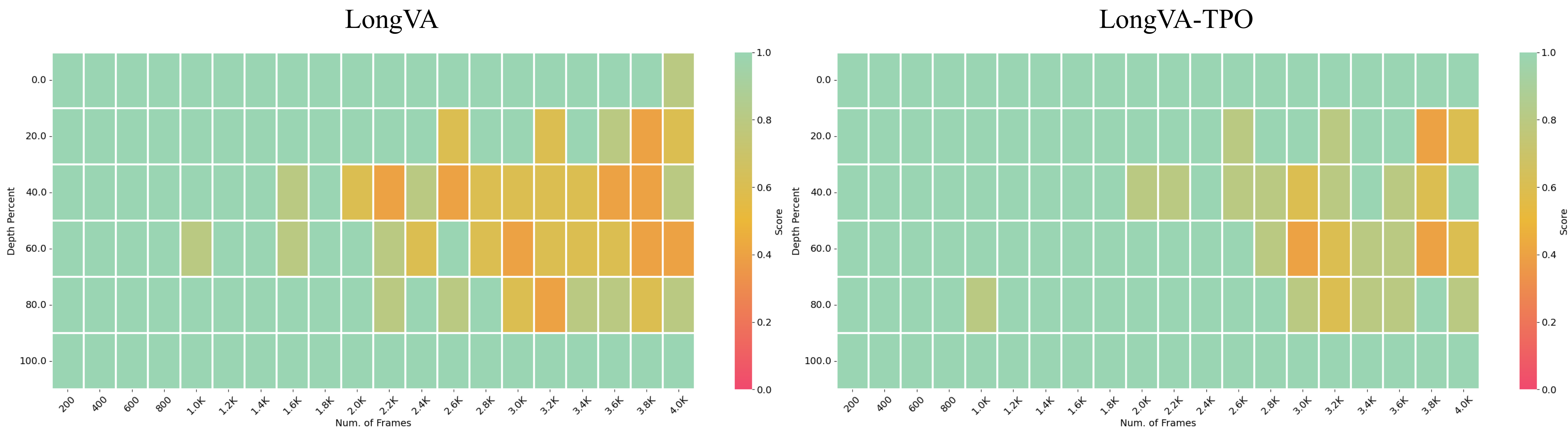}
    \end{minipage}

\caption{Performance comparison of LongVA and LongVA-TPO on the needle-in-a-haystack task across varying input video lengths (horizontal axis) and temporal depths (vertical axis). Heatmaps indicate improved temporal grounding capability of LongVA-TPO.}        \label{fig:needle}
\end{figure*}
\subsubsection{Effect of Dataset Sizes}

\begin{table}[h]
\centering
\begin{tabular}{p{1.5cm}ccc}
\toprule
\textbf{Model}  &\textbf{LongVideoBench}&\textbf{MLVU}& \textbf{VideoMME} \\
 \midrule%
LongVA&51.3&58.8&52.6 \\
TPO$_{\text{2k}}$& 52.5&57.8& 52.8\\
TPO$_{\text{5k}}$& 53.7 &59.5&53.6\\
TPO$_{\text{10k}}$&\textbf{54.2}&\textbf{61.7}&\textbf{55.1}  \\
\bottomrule
\end{tabular}
\caption{Results of LongVA-TPO (TPO) trained on different data scales. TPO achieves consistent performance improvements as the data scale increases. The performance on the VideoMME benchmark is evaluated without subtitles.}

\label{table:scale}
\end{table}
Scalability is a critical metric in the evaluation of algorithms in the era of large-scale models, reflecting an algorithm's performance as data volume expands. To examine the scalability of the TPO algorithm, we conduct experiments with LongVA-TPO across incremental sizes of 2k, 5k, and 10k (the complete preference dataset). The results, presented in Table \ref{table:scale}, highlight the impact of dataset scaling. Our findings reveal that LongVA-TPO demonstrates superior scalability, achieving consistent performance gains with increasing dataset size across all three benchmarks. This pattern highlights TPO’s robustness and adaptability in larger data contexts, suggesting its potential to deliver enhanced results when scaled to larger datasets.



\subsubsection{Effect of Post-Filtering}
As a critical component of the TPO framework, post-filtering effectively reduces noise and enhances data quality. To further assess its impact, we conducted experiments comparing the performance of LongVA-TPO with and without post-filtering. The results, presented in Table \ref{table:filtering}, demonstrate that post-filtering consistently improves performance across multiple benchmarks.
\begin{table}[ht]
\centering
\begin{tabular}{p{3cm}ccc}
\toprule
\textbf{Model}  &\textbf{LongVideoBench}&\textbf{MLVU}& \textbf{VideoMME} \\
 \midrule%
LongVA&51.3&58.8&52.6 \\
TPO$_{\text{w/o Post-Filtering}}$& 52.5&60.1&  53.6\\
TPO$_{\text{w/ Post-Filtering}}$&\textbf{54.2}&\textbf{61.7}&\textbf{55.1}  \\
\bottomrule
\end{tabular}

\caption{Results of LongVA-TPO (TPO) with and without post-filtering. Post-filtering consistently improves performance across multiple benchmarks.}
\label{table:filtering}
\end{table}

\begin{table}[ht]
\centering
\begin{tabular}{p{3cm}ccc}
\toprule
\textbf{ Ratio}  &\textbf{LongVideoBench}&\textbf{MLVU}& \textbf{VideoMME} \\
 \midrule%
10:0&53.5&58.7&54.0 \\
8:2& 53.8&59.9& 54.0\\
\textbf{5:5}~(\textbf{final~model})& \textbf{54.2}&\textbf{61.7}&\textbf{55.1}  \\
2:8&53.4&59.1&54.2\\
0:10&53.4&58.5& 53.8\\
\bottomrule
\end{tabular}
\caption{Performance of TPO across different training data mix ratios, varying the proportion of negative responses generated from incomplete versus irrelevant video segments.}

\label{table:ratio}
\end{table}
\subsubsection{Effect of Different Data Mix Ratio}
In TPO, we design two different kinds of manipulated schema for the dis-preferred response generation, including creating incomplete and irrelevant videos. 
To evaluate the individual capabilities, limitations, and necessity of combining incomplete and irrelevant videos, we conducted an ablation study. In this study, we maintained the same overall dataset size as the full TPO and assessed the performance of our TPO model under various data mixing ratios between the negative responses  generated from incomplete and irrelevant videos. 
The evaluated ratios included 10:0, 8:2, 5:5, 2:8, and 0:10. The experimental results, summarized in Table \ref{table:ratio}, clearly demonstrate that the model achieves optimal performance on general video understanding tasks when an equal proportion of  negative responses  generated from incomplete and irrelevant videos. This balanced data distribution effectively integrates different type of temporal information, leading to superior overall model performance.

\begin{figure}[t]
    \centering
    \includegraphics[width=0.5\linewidth]{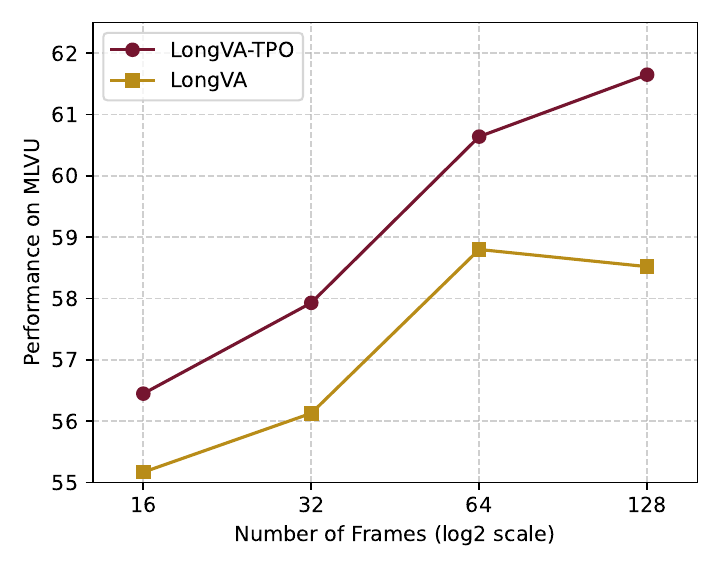}

\caption{Performance comparison between LongVA-TPO and LongVA on MLVU across varying input lengths. LongVA-TPO consistently benefits from increased input length, whereas LongVA's performance declines when inputs exceed 64 frames.}

    \label{fig:num_frame}
\end{figure}
\subsubsection{Needle-in-a-Haystack}

The Needle-in-a-Haystack task refers to the challenge of identifying a rare or specific event within a large volume of unstructured video data. Building on the work of \citet{zhang2024long}, we frame the task using image-based question answering (QA), where images are embedded within a 3-hour-long video, and the model is tasked with answering the corresponding image QA questions. In our experiments, we adopt the same five image QAs as~\citet{zhang2024long}, and present the results in Fig.~\ref{fig:needle}. While LongVA, optimized for long-context processing, significantly outperforms LLaVA-NeXT-Video \cite{zhang2024llavanextvideo} on the Needle-in-a-Haystack task (refer to Fig.~4 in~\cite{zhang2024long}), our LongVA-TPO model still demonstrates superior performance, achieving even better results in long-context temporal localization.

\subsection{Qualitative Analysis}
The qualitative analysis of our LongVA-TPO model and the LongVA model on two videos from the Video-MME benchmark is provided in Fig.~\ref{fig:qualitative}. In the first example, which involves a temporal localization and OCR task, our LongVA-TPO model demonstrates superior performance by accurately localizing the relevant information within the video and providing the correct answer to the OCR question. In the second example, a video discussing the Moon's formation, LongVA misinterprets the video content by relating it to the Earth's formation. In contrast, our LongVA-TPO model successfully comprehends and captures the key details of the video's content.


\section{Related Work}

\paragraph{Video Large Multimodal Models}
Recently, significant efforts have been devoted to extending the capabilities of large language models (LLMs) \cite{achiam2023gpt,reid2024gemini} into the visual domain,  developing various video large multimodal models (video-LMMs), including both proprietary \cite{achiam2023gpt,reid2024gemini} and open-source models \cite{Qwen2VL,liu2024visual,aria,shen2024longvu,lin2023video,chen2024internvl,yao2024minicpm,fu2025vita,abdin2024phi,laurençon2024matters}. Early approaches focused on curating high-quality video-text instruction-tuning datasets \cite{chen2024sharegpt4video,zhang2024videoinstructiontuningsynthetic,liu2023llava,park2023generative,zhang2024llavanextvideo}, to equip LLMs with video comprehension capabilities.  However, these datasets often rely on synthetic data derived from video captions, which limits their effectiveness in capturing visual-temporal dynamics. Other studies have focused on extending pretrained video-LMMs to handle longer video contexts \cite{zhang2024long,liu2024oryx,long-llava-qwen2-7b-2024,kangaroogroup,liu2024nvila,shu2024video,islam2025bimba}, while multimodal interleaved datasets \cite{li2024llavanext-interleave,lin2024vila} and mixed training strategies \cite{apollo,li2024llava} have been explored to enhance video-LMM performance.  Despite these advancements, the post-training stage for video-LMMs remains underexplored. Recent efforts like LLaVA-Hound \cite{zhang2024direct} utilize ChatGPT to rank model outputs and create preference datasets but fall short in leveraging the temporal information inherent in video data. In contrast, our work pioneers post-training strategies that explicitly incorporate temporal priors to address these limitations.

Temporal grounding is crucial for comprehending the video modality, particularly in long-form videos. Various efforts have been made to enhance temporal localization, including dense captioning \cite{wang2021end, yeung2018every, yang2023vid2seq}, highlight detection \cite{lei2021detecting, moon2023query}, and temporal video grounding \cite{gao2017tall, yuan2019semantic,xiao2024can}, among others. Recent advancements have introduced temporal-aware designs in video-LMMs~\cite{ren2024timechat, chen2024timemarker, li2023videochat, huang2024lita,wang2024grounded} and have explored the development of agentic systems with temporal grounding capabilities~\cite{wang2025videoagent}. Unlike these existing approaches, our work focuses on temporal preference optimization during the post-training stage, offering a complementary enhancement to current methods.




Proximal Policy Optimization (PPO) \cite{ouyang2022training, stiennon2020learning, ziegler2019fine} and Direct Preference Optimization (DPO) \cite{rafailov2024direct} are two widely used implementations of Reinforcement Learning from Human Feedback (RLHF) \cite{ouyang2022training, ziegler2019fine}, serving as key algorithms in preference learning and post-training. In the image-LMM domain, \citet{sun2023aligning} enhanced model visual capabilities by incorporating image captions into the reward modeling process within RLHF. Similarly, \citet{ahn2024tuning} fine-tuned multimodal foundation models using Reinforcement Learning from AI Feedback (RLAIF). Other approaches, such as those proposed by \citet{li2023silkie} and \citet{gunjal2024detecting}, directly distilled GPT-4V's preferences from sampled model responses. A notable strategy involves using text as an intermediate modality, leveraging captions and other descriptive information to extract and distill LLMs preferences for both images \cite{zhao2023beyond} and videos \cite{zhang2024direct}. Furthermore, \citet{pi2024strengthening}, \citet{zhou2024aligning}, and \citet{deng2024enhancing} advanced preference learning in image-LMMs by curating preference data through image input manipulation.

\paragraph{Self-Training in Foundation Models}
To address the challenge of scaling up annotated datasets, several works have explored self-improvement and self-training methods \cite{huang2022large,ho2022large}. \citet{zelikman2022star} introduced Self-Taught Reasoners (Star), which leverage generated chain-of-thought rationales to enhance LLMs' complex reasoning capabilities. In the image domain, BPO \cite{pi2024strengthening} STIC \cite{deng2024enhancing}  and POVID \cite{zhou2024aligning} improve image-LMMs responses by incorporating visual priors. In the video domain, Video-STaR \cite{zohar2024video} uses existing labels as weak supervision to guide model self-improvement while \citet{ahn2024srt} explores iterative self-improvement in preference optimization.

\section{Conclusion}
We introduced Temporal Preference Optimization (TPO), a scalable post-training framework that enhances temporal grounding in video-LMMs. By contrasting between the preference responses from the well-grounded and manipulated video clips, TPO effectively captures the intricate temporal dependencies required for video understanding. Extensive experiments across three challenging benchmarks—LongVideoBench, MLVU, and Video-MME—demonstrated TPO's robust improvements, achieving state-of-the-art performance. By integrating multi-granularity temporal preferences, TPO  offers a robust and efficient solution for advancing temporal reasoning in multimodal tasks.
One future direction is scaling the preference data to improve coverage and diversity, thereby enhancing TPO's generalizability. Additionally, while this work focuses on LongVA-7B and LLaVA-Video-7B as representative Video-LMMs, applying TPO to a broader range and larger scale of Video-LMMs would provide insights into its adaptability and performance across different architectures. 


{
    \small
    \bibliography{main}
}

\clearpage
\setcounter{page}{1}
\appendix

\section{Reproducibility Statement}
To ensure reproducibility, we have released the full scripts and source code of the TPO pipeline, accompanied by the curated preference dataset, which includes videos, associated queries, and corresponding preference responses, as well as the trained model weights. This release will include detailed implementations of all steps involved in the preference dataset curation and the preference optimization process. By providing these resources, we aim to facilitate the replication of our results and support further advancements in this area of research.

\section{Appendix overview}
This document provides more details of our approach and additional experimental results, organized as follows:
\begin{itemize}
	\vspace{2pt}
 	\item \S~\ref{implement} More Implementation Details of TPO.
	\item \S~\ref{dataset} More Details of the Preference Dataset Curation.
        \item \S~\ref{datasetexample} More Examples in the Preference Dataset.
 	\item \S~\ref{examples}  More Qualitative Examples.

\end{itemize}

{
\hypersetup{linkcolor=black} 
}

\section{Implementation Details}\label{implement}
We conduct Temporal Preference Optimization (TPO) on LongVA \cite{zhang2024long} and LLaVA-Video \cite{zhang2024videoinstructiontuningsynthetic}, two state-of-the-art  video-LMMs. The two TPO models are trained using 8 Nvidia A100 80GB GPUs, with a batch size of 64. For preference optimization, we set the KL-divergence weight ($\beta$) to 0.3 and the supervised fine-tuning (SFT) loss weight ($\alpha$) to 0.5 for LongVA-TPO and we set the KL-divergence weight ($\beta$) to 0.2 and the supervised fine-tuning (SFT) loss weight ($\alpha$) to 1 for LLaVA-Video-TPO. We employ full fine-tuning for both the multimodal projector and the language model while keeping the visual encoder frozen, using a learning rate of $4 \times 10^{-6}$ for LongVA-TPO and $3 \times 10^{-7}$ for LLaVA-Video-TPO. The training is performed on a curated dataset of 10k samples for one epoch for LongVA-TPO and 10k samples for one epoch for LLaVA-Video-TPO. A cosine learning rate scheduler with a warm-up ratio of 0.1 is utilized \cite{loshchilov2016sgdr}. The entire TPO fine-tuning process  takes approximately 4 hours on both two models.

For evaluation, we adopt the protocol outlined by LongVA \cite{zhang2024long} and LLaVA-Video \cite{zhang2024videoinstructiontuningsynthetic}, leveraging the official lmms-eval repository \cite{zhang2024lmmsevalrealitycheckevaluation} to assess our model's performance on three benchmarks. For LongVA-TPO, we set the parameter $max\_frames\_num=128$ across all three benchmarks. For LLaVA-Video-TPO, we set the parameter $max\_frames\_num=96$ for the Video-MME benchmark and  $max\_frames\_num=128$ for the rest of the benchmarks.

\section{Preference Dataset Curation}\label{dataset}
\begin{figure*}[htbp]
    \centering
    \begin{minipage}[t]{0.49\textwidth}
        \centering
        \includegraphics[width=\textwidth]{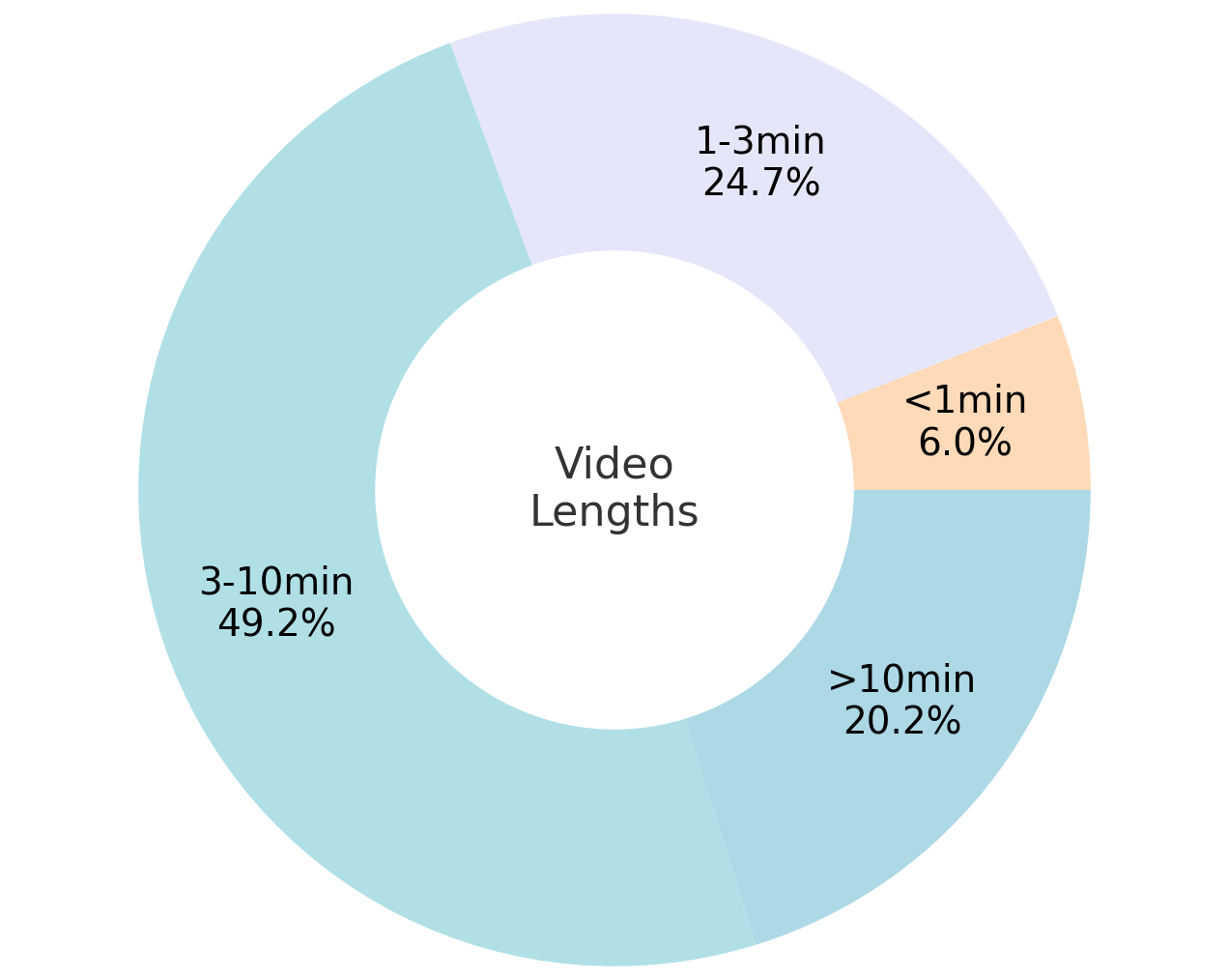}
        \caption{The distribution of lengths for 8K crawled videos.}
        \label{fig:length}
    \end{minipage}
    \hfill
    \begin{minipage}[t]{0.49 \textwidth}
        \centering
        \includegraphics[width=\textwidth]{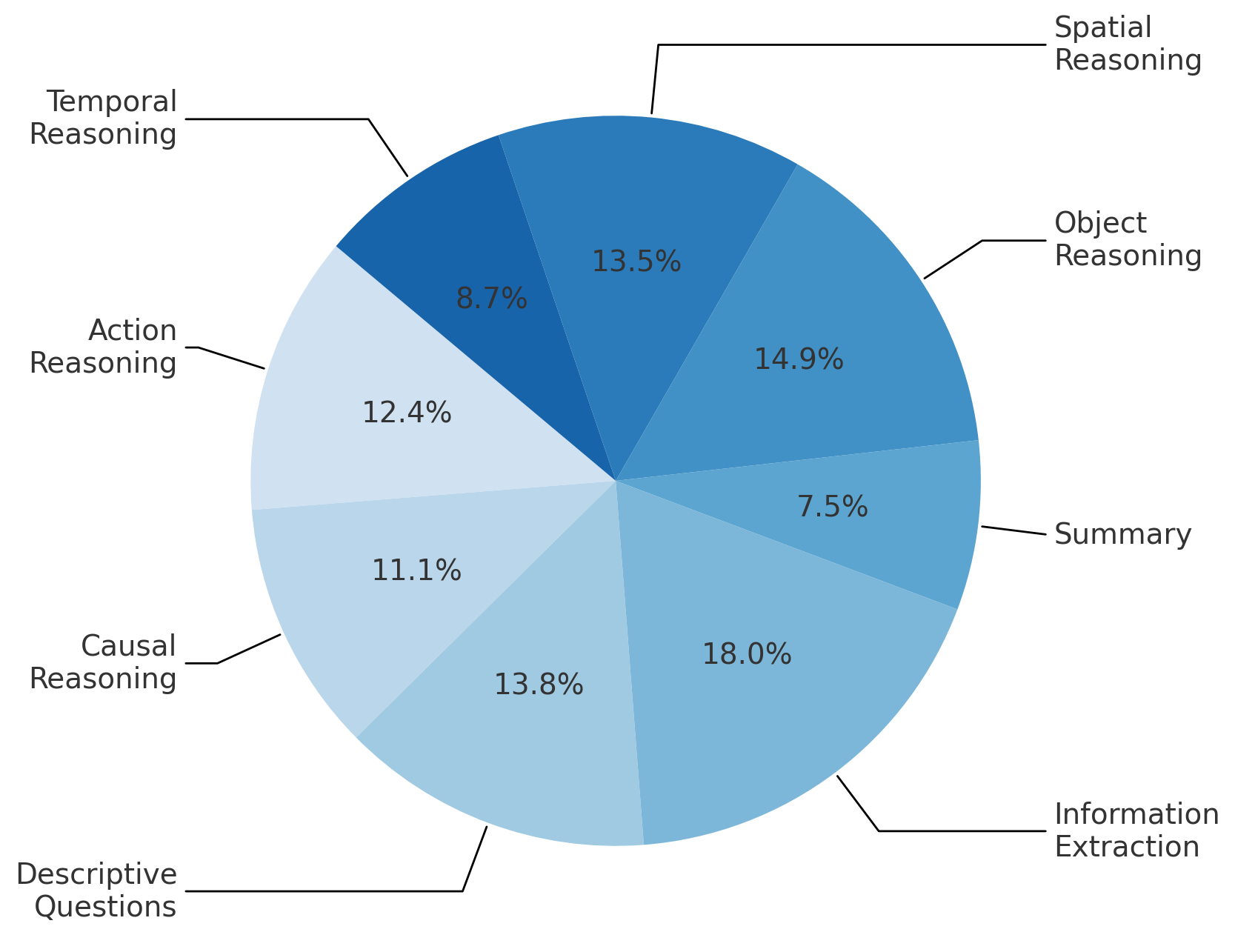}
         \caption{The distribution of question types for 10K curated preference dataset for LongVA-TPO.}
        \label{fig:type}
    \end{minipage}
    \vspace{0.1in}
    
\end{figure*}
We manually curated a set of 200 keywords assisted with GPT-4o-mini~\cite{achiam2023gpt}, which were utilized to retrieve 8,000 videos from the internet, forming a diverse and comprehensive dataset. Using this dataset, we further developed 10,000 queries paired with their corresponding preference responses, covering a broad range of tasks. The detailed prompts for preference dataset curation are provided in Fig.~\ref{fig:prompt-query} and Fig.~\ref{fig:prompt-filter1}. For LLaVA-Video, we  sampled a subset of 10k QA pairs from the LLaVA-Video-178k dataset with the negative responses only curated by incomplete videos.

The distribution of video lengths in our collected dataset is presented in Fig.~\ref{fig:length}. 
The distribution of tasks is illustrated in Fig.~\ref{fig:type}, encompassing Temporal Reasoning (8.7\%), Action Reasoning (12.4\%), Causal Reasoning (11.1\%), Information Extraction (18.0\%), Descriptive Questions (12.8\%), Summarization (7.5\%), Object Reasoning (14.9\%), and Spatial Reasoning (13.5\%).

\section{Preference Dataset Examples}\label{datasetexample}
We provide three additional examples of preference datasets, as illustrated in Fig.~\ref{fig:dataset_3}. For instance, in Example (a), the task involves an OCR-based query aimed at retrieving the quote located beneath a mural. The dis-preferred response incorrectly identifies the relevant frame, failing to locate the quote below the mural and instead referencing another frame containing the phrase ``Forward, Warrior.'' In contrast, the preferred response accurately identifies the corresponding frame based on the question. This is achieved by leveraging the highly relevant sub-video segment provided to the video-LMM, enabling the correct extraction of both the quote and its attribution.

For Example (b), the task involves summarizing information by identifying the four levels depicted in a pyramid diagram. The dis-preferred response, based on irrelevant video clips, provides incorrect names and an incorrect order for the four levels. In contrast, the preferred response accurately identifies both the correct names and the proper order of the four levels, demonstrating a better understanding of the context and alignment with the video content.

For Example (c), the task involves a high-level descriptive query requiring a summary of the exercise routine depicted in the video. The dis-preferred response, relying only on down-sampled frames, omits significant key information and provides an incomplete summary. In contrast, the preferred response accurately summarizes the entire exercise routine, offering both detailed and correctly ordered information, thereby demonstrating a comprehensive understanding of the video content.



\lstset{
  basicstyle=\ttfamily,
  keywordstyle=\color{blue},
  commentstyle=\color{green},
  stringstyle=\color{red},
  showstringspaces=false,
  breaklines=true,
  breakatwhitespace=true,
  postbreak=\mbox{\textcolor{red}{$\hookrightarrow$}\space}
}
\begin{figure*}[b]
\scriptsize
\begin{lstlisting}
<Video Caption>

Using the caption of a video, create a question-answer pair that focuses on <Task Prompt>.

Please generate a question tailored to the given caption. If it's inappropriate to generate such question, please output None.
Output format:
Q: <question>
A: <answer>

\end{lstlisting}
    \caption{
    \textbf{Detailed prompt for the query generation given the video captions.} 
    }
    \label{fig:prompt-query}
\end{figure*}

\lstset{
  basicstyle=\ttfamily,
  keywordstyle=\color{blue},
  commentstyle=\color{green},
  stringstyle=\color{red},
  showstringspaces=false,
  breaklines=true,
  breakatwhitespace=true,
  postbreak=\mbox{\textcolor{red}{$\hookrightarrow$}\space}
}
\begin{figure*}[b]
\scriptsize
\begin{lstlisting}
<Video Caption>

Question: <Query>
Answer1: <Preferred Answer>
Answer2: <Dis-Preferred Answer>

Task1: You are given a question, the golden answer and related captions. Is answer1 better than answer2?  Please answer with Yes or No or Equally.

Task2: Please check if this question and Answer1 contradicts to any part of the golden caption or this question might have another answer different from the given answer. Please respond with Yes or No.

Task3: Is the Answer1 is correct given the question and golden caption? Please respond with Yes or No.

\end{lstlisting}
    \caption{
    \textbf{Detailed prompt for the post-filtering process for the preference data pairs. } 
    }
    \label{fig:prompt-filter1}
\end{figure*}
\begin{figure*}[ht]
    \centering
    \includegraphics[width=0.88\textwidth]{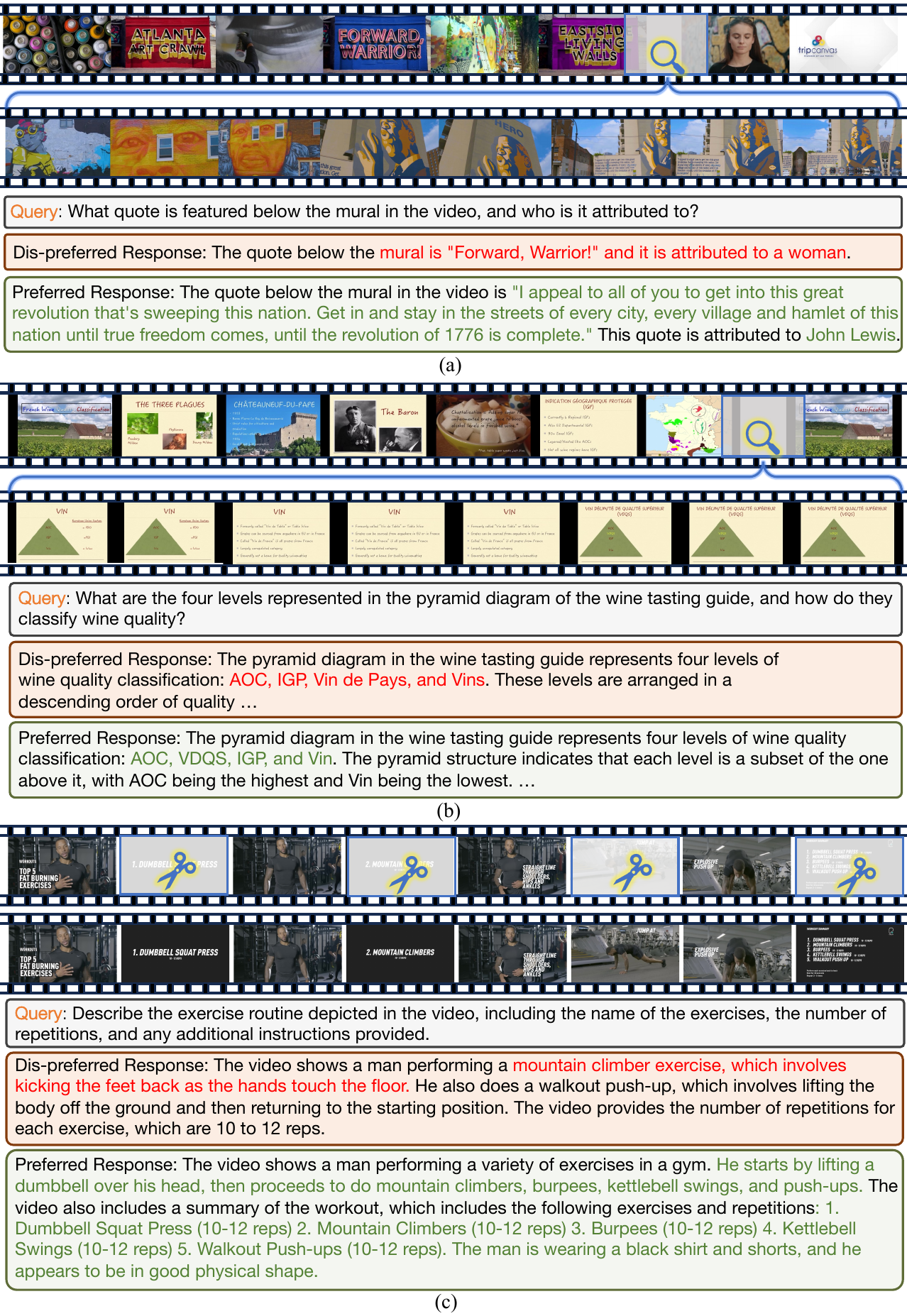}

    \caption{Examples from the preference dataset. 
    }
    \label{fig:dataset_3}
\end{figure*}

\section{Qualitative Analysis Examples}\label{examples}

\begin{figure*}[ht]
    \centering
    \includegraphics[width=\textwidth]{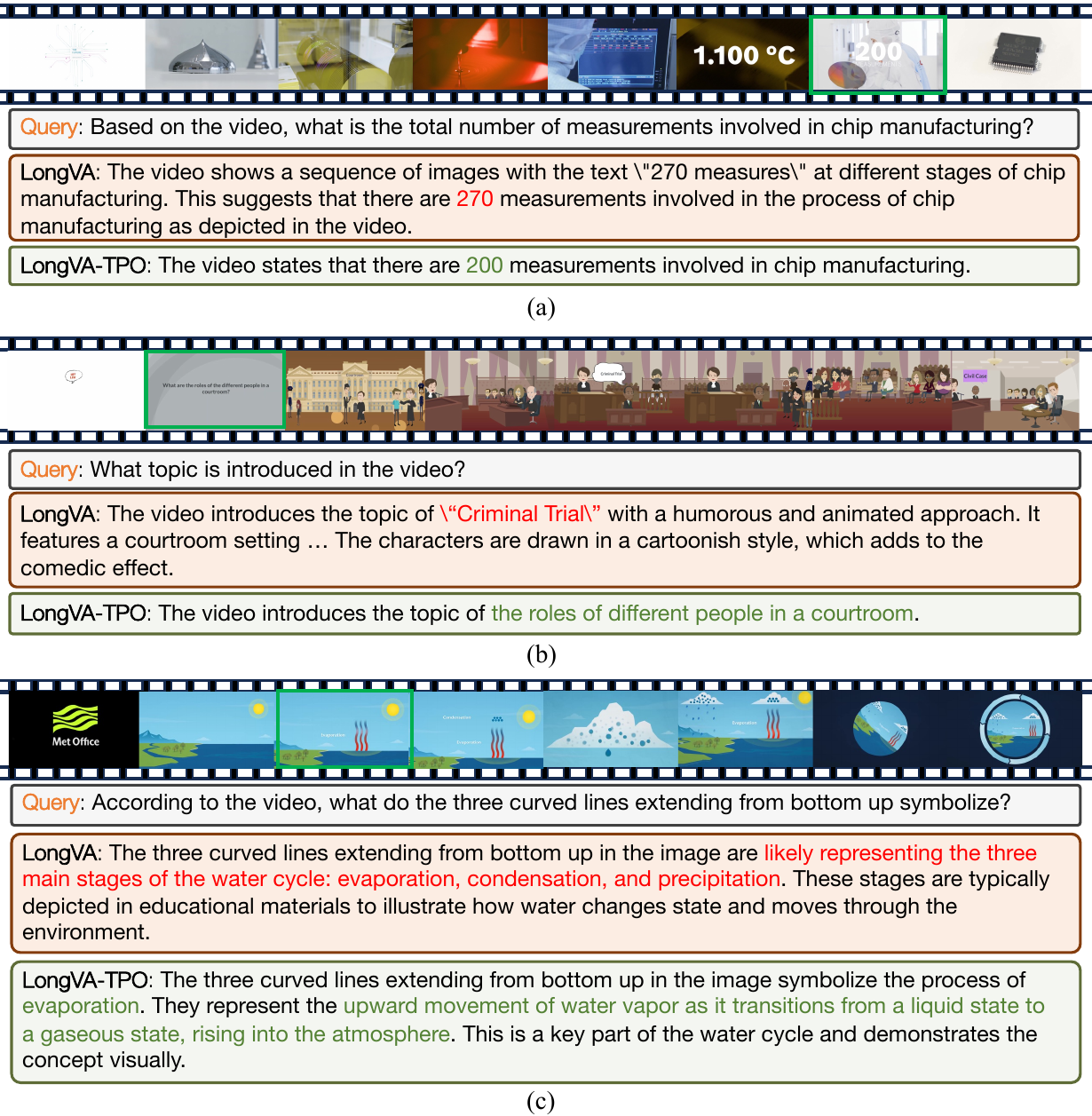}

    \caption{Additional qualitative examples from Video-MME \cite{fu2024video}. }

    \label{fig:qualitative_3}
\end{figure*}
We provide three additional qualitative analysis examples from the Video-MME dataset~\cite{fu2024video}, as illustrated in Fig.~\ref{fig:qualitative_3}. Example (a) involves an information extraction and optical character recognition (OCR) task, where the question asks for the total number of measurements involved in chip manufacturing. The original LongVA model failed to accurately locate the relevant frame containing the necessary information, resulting in an incorrect response. In contrast, our LongVA-TPO model, enhanced through temporal preference optimization, successfully identified the pertinent frame within the lengthy input and provided the correct answer to the question.

Example (b) involves a high-level video understanding and information extraction task, where the question asks for the main topic introduced in the video. The original LongVA model failed to capture the overarching theme, instead responding with an unrelated term, ``Criminal Trial,'' mentioned elsewhere in the video. In contrast, our LongVA-TPO model effectively identified the video's central theme and accurately provided the correct topic introduced in the content.

Example (c) involves an object reasoning task, where the question asks what the three curved lines extending from the bottom upward symbolize. The original LongVA model failed to interpret the representation accurately, erroneously stating that the lines represent three stages of the water cycle, which was a hallucination. In contrast, our LongVA-TPO model successfully understood the symbolic meaning of the three curved lines as representing evaporation, providing a correct and detailed response.

%

\end{document}